\documentclass[letterpaper, 10 pt, conference]{ieeeconf}
\IEEEoverridecommandlockouts 

\usepackage{amsmath}
\usepackage{amssymb}
\usepackage[nospace]{cite}
\usepackage{url}
\usepackage{xcolor}
\usepackage{fancyhdr}

\usepackage{adjustbox}
\usepackage{graphicx}
\usepackage{tensor}
\usepackage[utf8]{inputenc}
\usepackage[OT1]{fontenc}
\usepackage[final]{microtype}
\usepackage{booktabs}
\usepackage[binary-units]{siunitx}

\makeatletter
\let\NAT@parse\undefined
\makeatother
\usepackage[hidelinks]{hyperref}

\DeclareMathOperator*{\argmax}{arg\,max}

\newcommand{\SO}{\mathrm{SO}}
\newcommand{\SE}{\mathrm{SE}}
\newcommand{\SOthree}{\SO(3)}
\newcommand{\SEthree}{\SE(3)}
\newcommand{\Real}{\mathbb{R}}

\newcommand{\R}{\mathbf{R}}

\newcommand{\T}{\mathbf{T}}

\newcommand{\residual}{\mathbf{r}}

\newcommand{\tran}{\mathbf{p}}
\newcommand{\vel}{\mathbf{v}}
\newcommand{\bias}{\mathbf{b}}

\newcommand{\States}{\mathcal{X}}
\newcommand{\State}{\boldsymbol{x}}
\newcommand{\Measurements}{\mathcal{Z}}

\newcommand{\World}{\mathtt{W}}
\newcommand{\Odom}{\mathtt{O}}
\newcommand{\Base}{\mathtt{{B}}}
\newcommand{\Cabin}{\mathtt{{C}}}
\newcommand{\Lidar}{\mathtt{L}}
\newcommand{\Gnss}{\mathtt{G}}
\newcommand{\Imu}{\mathtt{I}}

\newcommand{\ba} {\bias^a} 
\newcommand{\bg} {\bias^g} 

\newcommand{\defeq}{\triangleq}

\IEEEoverridecommandlockouts
\title{\LARGE \bf

Graph-based Multi-sensor Fusion for Consistent Localization of Autonomous Construction Robots
}

\author{Julian Nubert$^{1,2}$, Shehryar Khattak$^{1,2}$ and Marco Hutter$^{1}$
\thanks{*This work is supported in part by the Max Planck ETH Center for Learning Systems, the EU Horizon 2020 programme grant agreement No.852044 and 101016970, the NCCR digital fabrication and robotics, and the SNSF project No.188596.}
\thanks{$^{1}$The authors are with the Robotic Systems Lab, ETH Z\"urich, {\tt\small\{nubertj, skhattak, mahutter\}@.ethz.ch}.}%
\thanks{$^{2}$The authors are with the Max Planck ETH Center for Learning Systems.}%
\thanks{Corresponding Author: Julian Nubert, \tt\small nubertj@ethz.ch}%
}

\begin{document}

\maketitle
\thispagestyle{empty}
\pagestyle{empty}

\begin{abstract}
Enabling autonomous operation of large-scale construction machines, such as excavators, can bring key benefits for human safety and operational opportunities for applications in dangerous and hazardous environments. To facilitate robot autonomy, robust and accurate state-estimation remains a core component to enable these machines for operation in a diverse set of complex environments. 
In this work, a method for multi-modal sensor fusion for robot state-estimation and localization 
is presented, enabling operation of construction robots in real-world scenarios. The proposed approach presents a graph-based prediction-update loop that combines the benefits of filtering and smoothing in order to provide consistent state estimates at high update rate, while maintaining accurate global localization for large-scale earth-moving excavators. 
Furthermore, the proposed approach enables a flexible integration of asynchronous sensor measurements and provides consistent pose estimates even during phases of sensor dropout. For this purpose, a dual-graph design for switching between two distinct optimization problems is proposed, directly addressing temporary failure and the subsequent return of global position estimates.
The proposed approach is implemented on-board two Menzi Muck walking excavators and validated during real-world tests conducted in representative operational environments. 
\end{abstract}

\section{Introduction}\label{sec:intro}
Large-scale construction excavators represent one of the most versatile and powerful class of equipment due to their capabilities in a wide range of real-world tasks. Given their ability to move heavy loads, operate for long periods and traverse over most challenging terrains, they play a key role in many essential industries, such as construction, material handling, and forestry. In addition to industrial applications, these machines are also utilized in disaster response operations and undertake heavy duty tasks such as the removal of earthquake debris, the disposal of hazardous materials, and the creation of trenches to divert flood waters. However, despite being directly involved in many critical tasks, practical applications are still fully reliant on a human operator. This dependence, in addition to putting humans in harm's way, also restricts the type of tasks and the range of operational conditions. Although efforts are made to improve the state of autonomy, practical applications still remain limited to certain predefined tasks.

\pdfpxdimen=\dimexpr 1 in/72\relax
\begin{figure}[ht!]
\centering
    \adjustbox{trim=0 0 0 0,clip, width=\columnwidth}
    {\includegraphics{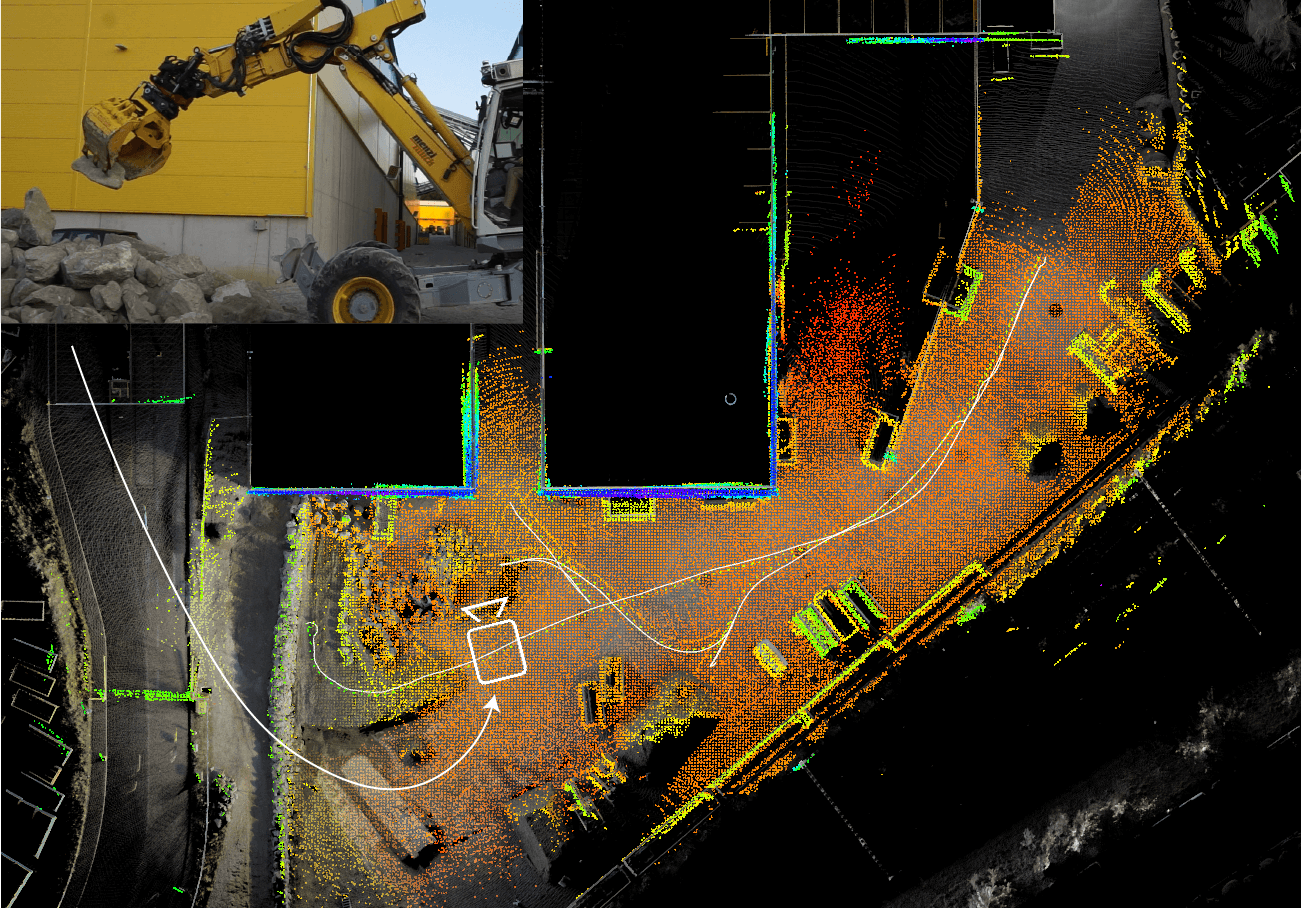}}
\caption{A global map built online using the proposed approach's estimate during a real-world construction task is shown. The colored robot map is overlaid on top of a groundtruth map created using a Leica RTC360 scanner.}
\label{fig:main}
\vspace{-2ex}
\end{figure}

Towards the goal of enabling autonomy of these machines, one core component that requires attention is robust and reliable state-estimation and localization for the described scenarios. Smooth and consistent state-estimates at high frequency are required for control tasks such as chassis balancing~\cite{hutter2016force}. Similarly, to facilitate high-level robotic autonomy tasks such as navigation~\cite{jelavic2021forest}, planning~\cite{GBPlanner} and mapping~\cite{heap2021stones}, accurate localization is required.  
Given the diversity and complexity of the operational environments for these machines, it is essential to develop robust yet flexible pose estimation approaches to handle special events such as sensor loss.

Motivated by the aforementioned challenges, this work presents a robust and flexible multi-modal sensor-fusion approach that is capable of providing low-latency state-estimates while remaining globally accurate. In particular, an asynchronous fusion of IMU, lidar and Global Navigation Satellite System (GNSS) estimates is proposed within a flexible framework. To achieve high update rates a graph-based prediction-update loop is proposed that aims to combine the advantages of filtering and optimization-based approaches. Furthermore, this work proposes the concurrent usage of multiple factor graphs to maintain state-estimation consistency even when localization undergoes large global corrections. The proposed approach is thoroughly evaluated during real-world field deployments (cf. Figure~\ref{fig:main}) onboard two different excavators, demonstrating that consistent and accurate pose estimates are provided even for events of sensor loss and recovery. Finally, the code of the proposed approach is made publicly available for the benefit of the robotics community\footnote{\label{note:github} \url{https://github.com/leggedrobotics/GMFCL}}.

\section{Related Work}\label{sec:related}
Reliable state-estimation and accurate localization are widely studied topics in the field of robotics. A number of filtering~\cite{lynen13robust,bloesch2017two}, smoothing~\cite{indelman2013information,kilic2019improved} and batch-optimization~\cite{diehl2009efficient,sandy2019confusion} based methods have been proposed to provide smooth and consistent robot state-estimates at low-latency for control tasks. Similarly, a number of approaches utilizing a diverse set of sensors have been proposed~\cite{loam2018, geneva2020openvins, nubert2021self, hong2020radarslam, imuGPS2020} to enable accurate robot localization at large-sale and to facilitate high-level tasks such as planning, navigation and mapping.

However, for the automation of large earth moving machines few efforts have been made to enable robust pose estimation for operation in challenging environments. Current works, such as ~\cite{mineWallManipulator2019,ROBDEKON}, have tried to directly apply existing methods on construction machines to enable automation for specific tasks, however, for general applications recent works have emphasized on the design of specific and flexible approaches to enable general autonomy for excavators~\cite{militaryConstruction2019}. In particular,~\cite{ardiny2015review,petersen2019review} present a comprehensive review of the state-of-the-art in constructions robotics and stress on the utilization of multi-sensor perception and fusion to enable operation in complex environments. Generally, as construction work using large machines takes place outdoors, accurate external positioning from GNSS is typically available. Taking advantage of this, the walking excavator HEAP (Hydraulic Excavator for an Autonomous Purpose)~\cite{heap2021} - which is also deployed in this work - has been developed, and which utilizes IMU and RTK-GNSS fusion for state-estimation. With the current approach the excavator is capable of performing construction tasks autonomously, however, operation in GPS-denied environments remains a challenge. Citing similar concerns,~\cite{markerPoseExcavaor2021} propose to place artificial markers on the manipulator and track it using external cameras for excavation tasks. Although low-cost, such an approach is only suitable for tasks where the excavator is either static or operates within a limited workspace as markers need to remain within the detection range of the cameras. To increase the excavator's workspace,~\cite{extCamLocForConstruction2020} and ~\cite{stereoGPSfusion2018} employ deep-learning and stereo-GPS fusion techniques, respectively, to remove the need for artificial markers. However, installation of external cameras on work sites is still required, hence, limiting the operation of excavators to predefined areas.

Self-sufficient and autonomous operation in diverse environments for large-scale applications and over longer periods of time requires integration and utilization of onboard sensing as shown by ~\cite{autonomousExcavator2021} during continuous autonomous operation of an excavator for 24 hours. The proposed system utilized GPS for global positioning and a 3D lidar to refine excavator pose against a map of the environment. Similarly, another work~\cite{jelavic2021forest} demonstrated the use of an excavator for forestry applications, however, to enable under canopy operations a human operator had to manually map the forest plot prior to excavator deployment.
Furthermore,~\cite{heap2021stones} used aerial and ground maps with GPS-IMU odometry estimation to refine the robot pose for building a dry-stack wall.

Motivated by the discussion above, this work proposes a multi-modal sensor fusion approach that uses a multi-threaded architecture to employ multiple factor graphs~\cite{dellaert2017factor} in parallel. This design allows the approach to provide consistent low-latency state-estimates while remaining globally accurate as part of a single solution. 

\section{Problem Formulation}\label{sec:approach}
The formulation of the state-estimation and localization problem for a walking excavator is presented in this section.

\subsection{Frame Definitions}
The used reference frames are defined as: the fixed-world frame $(\World)$, the local odometry frame $(\Odom)$, the IMU frame $(\Imu)$, the lidar frame $(\Lidar)$, and the GNSS frame $(\Gnss)$.
Furthermore, the excavator chassis base frame $(\Base)$ and cabin frame $(\Cabin)$ are defined, as they are required for generating driving motions and controlling the chassis~\cite{hutter2016force}, the control of the arm~\cite{egli2021general} and the cabin~\cite{jud2021robotic}, respectively. These two frames are rotated against each other through the cabin turn joint. Throughout this work, the sensor extrinsic calibration is assumed to be available. Refer to~\cite{heap2021} for more details on the excavator's kinematic structure. 

\subsection{Definition of the State}
The excavator state at time $t_i$ consists of
\begin{equation}
    \State_i \triangleq \left[^\Imu \State_i, \T_{i\World \Odom}, ^\Base \State_i\right],
    \label{eq:state-definition}
    \vspace{-2mm}
\end{equation}
with
\vspace{-2mm}
\begin{equation}
^\Imu \State \triangleq \left[\R_{\World \Imu},\tensor[_\World]{\tran}{_{\World \Imu}},\tensor[_\World]{\vel}{_{\World \Imu}},\tensor[_\Imu]{\bias}{^g},\tensor[_\Imu]{\bias}{^a} \right] \in \SOthree \times \Real^{12}
\end{equation}
being the state of the IMU with respect to $\World$, $\T_{\World \Odom} \in \SEthree$ being the transformation from $\Odom\rightarrow\World$, and 
\begin{equation}
^\Base \State \triangleq \left[\R_{\Odom \Base},\tensor[_\Odom]{\tran}{_{\Odom \Base}},\tensor[_\Odom]{\vel}{_{\Odom \Base}}\right] \in \SOthree \times \Real^{6}
\end{equation}
denoting the state of $\Base$ w.r.t $\Odom$.
Here $\R \in \SOthree$ is the orientation, $\tran \in \Real^{3}$ is the position, $\vel \in \Real^{3}$ is the linear velocity, and $\bg \in \Real^{3}$ and $\ba \in \Real^{3}$ are the IMU biases.
The set of past and current IMU states in $\World$ up to time $t_k$ is denoted as
$^\Imu \States_k \dot{=} \{^\Imu \boldsymbol{x}_i\}_{i \in _\Imu\mathcal{K}_k}$, with $\mathcal{K}_k$ being introduced in the following subsection.

\begin{figure*}[ht!]
\includegraphics[width=\textwidth]{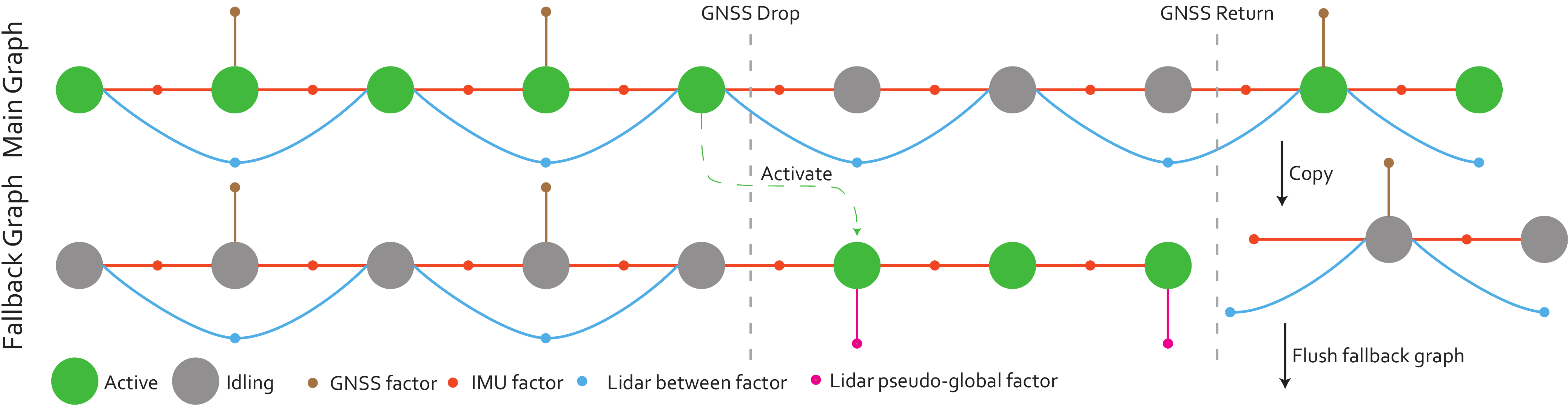}
\caption{Structural overview of the factor graph design proposed in this work. As depicted in this figure, IMU measurements are the driving factor in the architecture. Furthermore, a dual graph design is introduced in order to remain operational also in absence of RTK GNSS.}
\label{fig:graph_overview}
\vspace{-2ex}
\end{figure*}

\subsection{Measurements}
The base and cabin IMU measurements at time $t_k$ are denoted as $_\Base \mathcal{I}_k$ and $_\Imu \mathcal{I}_k$, respectively. Both include linear acceleration $\Tilde{\mathbf{a}}$ and angular velocity $\Tilde{\boldsymbol{\omega}}$ measurements. In order to connect the base to the cabin frame, discrete angle measurements of the cabin rotation $\mathcal{M}_i$ are provided. Next, the GNSS frame positions are provided in the global spherical coordinates but then converted into Cartesian world frame measurements, denoted as $_\World \mathcal{G}_i$. Lastly, lidar measurements in $(\Lidar)$ are given as $\mathcal{_\Lidar \mathcal{L}}_i$. Hence, the set of measurements until time $t_k$ can be written as $\Measurements_k \defeq \lbrace ^\Imu \Measurements_k, ^\Base \Measurements_k\rbrace$, with
\begin{equation}
    ^\Imu \Measurements_k \defeq \lbrace _\Imu\mathcal{I}_i, _\World\mathcal{G}_j, _\Lidar \mathcal{L}_m \rbrace_{i \in _\Imu\mathcal{K},j \in _\Gnss\mathcal{K},m \in _\Lidar\mathcal{K}},~\text{and}
\end{equation}
\begin{equation}
    ^\Base \Measurements_k \defeq \lbrace _\Base \mathcal{I}_i, \mathcal{K}_i \rbrace_{i \in _\Base\mathcal{K}}.
\end{equation}
The sets of all IMU, lidar and GNSS measurements that have arrived until time $t_k$ are denoted as $_\Imu \mathcal{K}_k$, $_\Lidar \mathcal{K}_k$, and  $_\Gnss \mathcal{K}_k$, respectively, and $\mathcal{K}_i$ denotes the measured joint positions..

\subsection{Objective}
The problem objective at time $t_k$ consists of three parts:
\begin{enumerate}
    \item Maximum a posteriori estimation for all past IMU states given all provided measurements, i.e.
        \begin{equation}
            ^\Imu \States_{i}^\star = \argmax_{^\Imu \States_i} p(^\Imu  \States_i|^\Imu \Measurements_i) \propto p(^\Imu \States_0)p(^\Imu \Measurements_i|^\Imu  \States_i)
            \label{equ:MaxAPost}
        \end{equation}
    \item Determining an accurate and locally consistent estimate of the robot state in $\Odom$ by maintaining $\T_{\World \Odom}$ transform.
    \item Obtaining a smooth state estimate $^\Base \State$ to facilitate robot control.
\end{enumerate}
Given a locally consistent estimate of $^\Imu \boldsymbol{x}$ and $\T_{\World \Odom}$, $^\Base \boldsymbol{x}$ can be computed using kinematics and the base IMU measurements $_\Base \mathcal{I}$. Hence, this work presents solutions for 1) and 2).

\section{Proposed Approach}\label{sec:approach}
In this section the proposed solution for obtaining consistent state estimates at a high update rate while maintaining accurate global localization for construction tasks is outlined.

\subsection{State Estimation using Factor Graphs}
This subsection formulates the mathematical components of the objective defined in Equation~\ref{equ:MaxAPost} more precisely, and introduces the deployed graph factors and error functions.

\subsubsection{Maximum a Posteriori Estimation}
\label{sec:maxAPost}
To obtain the optimal state estimate $^\Imu \State_k$ at the current time $t_k$, Equation~\ref{equ:MaxAPost} needs to be solved.
By iteratively applying Bayes' rule to the states within the considered optimization horizon, the posterior distribution of $^\Imu \States_k$ can be factorized into prior and likelihood terms. With the assumption of Gaussian error for each of the measurements $^\Imu \boldsymbol{z}_i \in~^\Imu \Measurements_k \sim \mathcal{N}(h(^\Imu \State_i),\,\sigma_k^{2})$ compared to the generally nonlinear measurement function $h(^\Imu \State_i)$, the problem can be rewritten as a least squares optimization with residuals $\boldsymbol{r}_{^\Imu \boldsymbol{z}_{k}} = h(^\Imu \boldsymbol{x}_k) - ^\Imu \boldsymbol{z}_k$. Factor graphs~\cite{dellaert2017factor} are a convenient way to formulate such optimization problems in the form of bipartite graphs consisting of variables and constraints.
In the presented solution the state estimation takes place in the IMU frame. An overview of the proposed approach is shown in Figure~\ref{fig:graph_overview}. The dual factor graph design is introduced later in this section.

\subsubsection{IMU Factors}
For each arriving IMU measurement a new state variable node is created in the graph. As IMU readings are received at a significantly higher rate than other sensor measurements, the created nodes serve as anchor points for other measurements. Each new node added to the graph is connected to past nodes using an \emph{IMU factor}, which is modeled as introduced in~\cite{forster2016manifold} and defines error terms as:
\begin{equation}
\sum_{i\in _\Imu\mathcal{K}_k} \left( \|\residual_{_\Imu\mathcal{I}_i}\|^2_{\Sigma_{_\Imu\mathcal{I}_i}} \right),~\text{with}~
\residual_{\mathcal{I}_i} \dot{=} \left[\residual_{\Delta\R_i}^\top, \residual_{\Delta v_i}^\top, \residual_{\Delta p_i}^\top \right],
\end{equation}
with covariance $\Sigma_{_\Imu\mathcal{I}_i}$ defined in Equation (45) of \cite{forster2016manifold}.

\subsubsection{GNSS Measurement}

If RTK GNSS is available, global position estimates with centimeter level accuracy can be obtained~\cite{maddern2020real}. This information is integrated into the optimization in the form of a GNSS factor, which includes a position residual
\begin{equation}
    \sum_{i\in _\Gnss\mathcal{K}_k} \left( \|\residual_{_\Gnss\mathcal{G}_i}\|^2_{\Sigma_{_\Imu\mathcal{I}_i}} \right),~\text{with}~\residual_{_\Gnss\mathcal{G}_i}= \tensor[_\World]{\tran}{_{\World \Imu}} - \tensor[_\World]{\Tilde{\tran}}{_{\World \Imu}},
\end{equation}
with $\tensor[_\World]{\Tilde{\tran}}{_{\World \Imu}}$ being the measured quantity. GNSS position measurements need to be expressed as IMU positions in the world frame before they can be added to the factor graph and hence they are first transformed as:
\begin{equation}
    \tensor[_\World]{\Tilde{\tran}}{_{\World \Imu}} = \tensor[_\World]{\Tilde{\tran}}{_{\World \Gnss}} + \R_{\World\Gnss} \cdot \tensor[_\Gnss]{\tran}{_{\Gnss\Imu}}.
\end{equation}

\subsubsection{Lidar Measurements}
Lidar odometry measurements are transformations between consecutive lidar scans computed using the approach described in CompSLAM~\cite{khattak2020complementary}.
For different scenarios lidar measurements are expressed with two types of factors (cf. Section~\ref{sec:dual_graph_opt}).
\paragraph{Lidar Odometry Factor}
As shown in Figure~\ref{fig:graph_overview}, during global operation the lidar measurements are directly added as a 6 DOF odometry factor. The difference between the predicted relative transformation of the odometry and the measurement model $^\Imu h(\boldsymbol{x})$ is evaluated on the underlying manifold. Note that this factor is required in order to render the global yaw observable in presence of GNSS position.
\paragraph{Lidar Pose Unary Factor}
In absence of GNSS measurements, the lidar measurements are transcribed into a pseudo global that are then used as unary pose factors, expressed in $\Imu$, in order to make use of the mapping capabilities of CompSLAM, and to avoid accumulated drift when only relative measurements are used. 
In order to benefit from roll and pitch estimates from IMU, due to observability, higher covariances are set for roll and pitch estimates from lidar odometry. Similar as for the lidar odometry factor, the error of $\T_{\World\Imu}$ against the measurement model $^\Imu h(\boldsymbol{x})$ is evaluated on the underlying manifold.

\subsubsection{Initialization}

For initialization the construction machine remains static for few seconds.
This allows for the accumulation of IMU readings and GNSS measurements from both antennas. These accumulated measurements as well as the heading obtained from the two antennas are then used to estimate the initial attitude $\R_{\World\Imu}$, and IMU biases ($\bg$,$\ba$). 
Given the IMU heading in the world, the position of the left GNSS antenna is used to compute the global initial position $\tensor[_\World]{\tran}{_{\World \Imu}}$. Yaw initialization significantly improves the convergence time of the optimizer at start up as compared to starting with an arbitrary initial yaw estimate.

\subsection{Prediction-Update Loop for Fast State Estimation}
\label{sec:prediction_update}
As motivated earlier, state estimates at a high update rate are required for practical control tasks. Typically, filtering-based approaches are employed due to their lightweight nature, however, they are limited in their ability to deal with highly delayed and nonlinear measurements. Optimization-based methods are more suitable for handling delayed and asynchronous measurements, however, for state-estimation problems solving a comprehensive non-linear optimization problem within the range of a milliseconds is still challenging.

To combine the benefits of filtering and optimization-based methods, this work proposes a novel prediction-update loop design for factor graphs.
The proposed algorithm is designed to take advantage of the mutli-threaded capabilities of modern CPUs. Each sensor measurement is handled and added to the optimization in a separate thread. Besides this, the optimization thread is running, awaiting the trigger being released by either GNSS or lidar callbacks. Upon arrival of an IMU measurement, only $\R_{\World \Imu}$,$\tensor[_\World]{\tran}{_{\World \Imu}}$ and $\tensor[_\World]{\vel}{_{\World \Imu}}$ are propagated and the IMU mesaurement is added to the factor graph without triggering the optimization. This allows for fast state-estimation suitable for controls tasks.  
The update thread optimizes the full receding-horizon factor graph and updates the full state estimate $^\Imu \State$.
Before running the optimization the newly arrived factors are copied over into the graph to be optimized. As soon as the result for the optimization is complete, the last optimized state is propagated to the latest IMU state using the obtained IMU biases, acting similarly to an update step in filtering-based approaches. Note, that this graph-based approach inherently allows to integrate delayed measurements into the smoothing operation at each time-step.

\subsection{Dual Factor Graph Design}
\label{sec:dual_graph_opt}

The approach introduced in the last section is able to obtain fast and smooth updates, however, this assumes that all measurements are free of outliers. This section outlines a solution based on a dual factor graph scheme (Figure~\ref{fig:graph_overview}) for the detection and handling of sensor loss and recovery, e.g. intermittent GNSS outage. Furthermore, measurement estimates can become unreliable due to sensor data degradation. An example is given by RTK GNSS measurements for which tall structures (e.g. walls), covered areas (e.g. under trees or in a tunnel), or internet outage usually leads to heavy degradation of the performance or even complete system failure.

\subsubsection{Outlier Rejection}
To detect and suppress GNSS outlier measurements, covariance estimates directly provided by modern RTK GNSS sensors are used. Furthermore, measurement inconsistency is detected by only allowing delta transformations up to a certain threshold (defined by the maximum velocity). As both criteria need to be satisfied several time steps in a row, outliers can be detected reliably.

As depicted in Figure~\ref{fig:graph_overview}, two parallel factor graphs are deployed. When reliable GNSS estimates are available the main graph is used, which includes lidar measurements as relative odometry factors. In this scenario the GNSS factors keep the estimate globally accurate. If GNSS signal becomes unreliable, a switch to the fallback graph is performed in order to maintain high accuracy in the world frame. At that point of switch both graphs are identical, however, in the fallback graph the lidar measurements are now added as \emph{pseudo-global unary factors} to maintain consistency with the onboard localization and mapping solution. 
Please note that during operation in fallback graph, relative lidar odometry factors are still being added to the main graph, but its output is not used for state-estimation or localization purposes.

Let the time of the GNSS loss be denoted as $t_j$. Then the pseudo global factor at time $t_k$ is computed as
\begin{equation}
    \Tilde{\T}_{\World\Imu_{t_k}} = \T_{\World\Imu_{t_j}} (\T_{\Lidar_{t_0}\Lidar_{t_j}})^{-1} \T_{\Lidar_{t_0}\Lidar_{t_k}},
\end{equation}
where $\T_{\World\Imu_{t_j}}$ being the last global estimate before GNSS loss, $\T_{\Lidar_{t_0}\Lidar_{t_j}}$ being the measurement in the lidar odometry frame at time of GNSS loss, and $\T_{\Lidar_{t_0}\Lidar_{t_k}}$ being the measurement in the lidar odometry frame at the current time.

\subsubsection{Localization Inconsistency on GNSS Recovery}
While the recovery of GNSS measurements is beneficial for correction of global localization drift, this however can cause non-smooth jumps in the localization and can be problematic for control tasks if not handled properly. Upon GNSS recovery a switch back to the main graph is performed which can lead to large jumps in position due to accumulated drift from the IMU and relative lidar odometry factors.
Yet, as the time of GNSS recovery is known, the relative drift between world and odometry frames can be estimated, and $\T_{\World\Odom}$ updated by:
\begin{equation}
    \T_{\World\Odom} = \T_{\World\Imu} \cdot (\T_{\Odom\Imu})^{-1},
\end{equation}
where $\T_{\Odom\Imu}$ is the estimate in the odometry frame of the fallback graph, whereas $\T_{\World\Imu}$ is already provided by the main graph.
Finally, the fallback graph is reset to the state of the main graph upon GNSS recovery.

\subsection{Implementation Details}
\label{sec:implementation_details}
The presented architecture, including the prediction-update loop and the dual factor graph design is implemented in C++ using the GTSAM framework~\cite{dellaert2012factor}. 
The lidar scans are obtained from an Ouster OS0-128, with odometry measurements provided by CompSLAM~\cite{khattak2020complementary}, however, pose estimates from other frameworks could be utilized as well. 
The GNSS estimates are provided by a \textit{Leica iCON iXE3} GNSS system with two antennas. For the latter, RTK corrections are received from permanently installed base stations. The system detects when GNSS estimates become less reliable and increases the published covariance internally. 
The proposed framework will be made publicly available in a generalized and easy-to-use form upon publication of the paper\textsuperscript{\ref{note:github}}.

\section{Experimental Results}\label{sec:evaluation}
In this section two experiments are presented: i) a construction operation showcase under challenging environmental conditions in Oberglatt, Switzerland, and ii) navigation in a more urban environment in Wangen, Switzerland. Within this section, the first experiment is referred to as \emph{Construction Task}, while the second will be denoted \emph{Navigation task}. An illustration of the first showcase and its corresponding groundtruth map are shown in Figure~\ref{fig:main}. 
\begin{figure}[t]
\includegraphics[width=\columnwidth]{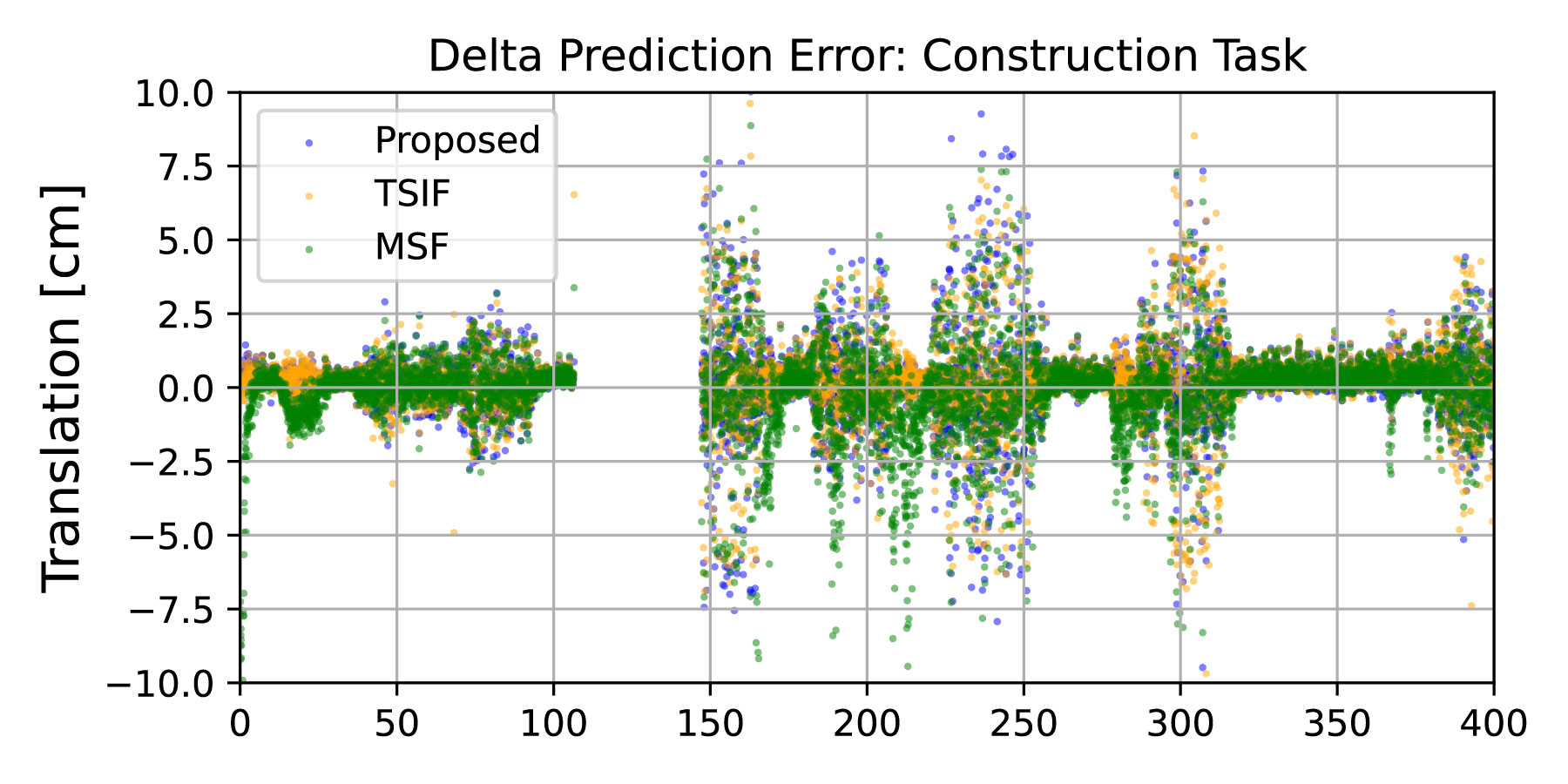}
\includegraphics[width=\columnwidth]{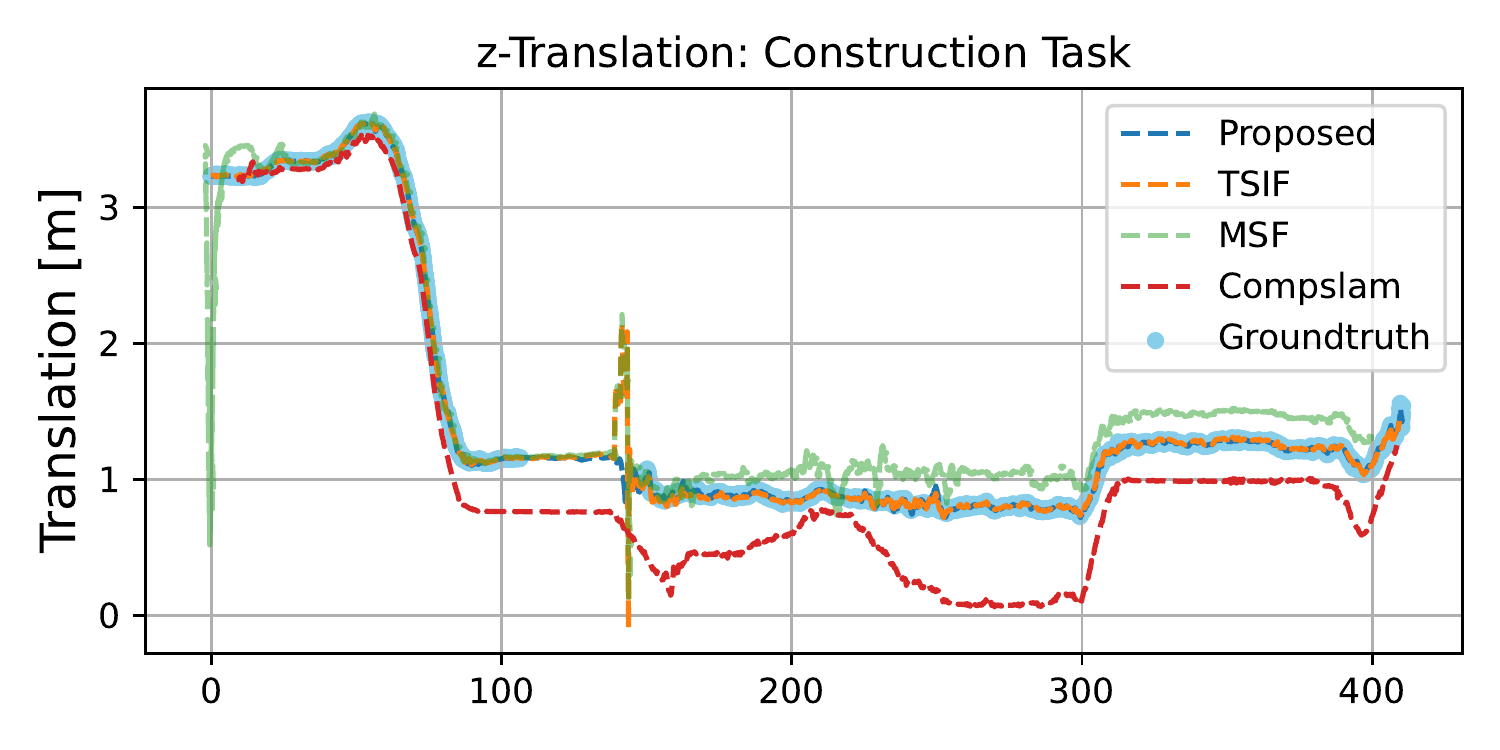}
\caption{Global accuracy for the real-world construction task. The top plot shows the relative position errors with respect to GNSS measurements, whereas the lower plot underlines the benefit of the dual-graph design in absence of GNSS from second $105$ to $150$. Time is expressed in [s].}
\vspace{-2ex}
\label{fig:accuracy}
\end{figure}
A thorough comparison is performed for both scenarios against the previously deployed solution on the HEAP~\cite{heap2021} excavator - namely the Two-State implicit filter (TSIF)~\cite{bloesch2017two} - as well as against an open-source multi-sensor fusion approach, MSF, based on an extended Kalman filter~\cite{lynen13robust}.
The first one, referred to as \emph{TSIF} within this section, performs state estimation based on IMU readings and GNSS measurements. In order to also compare against a framework fusing all three sensor modalities, results obtained from \emph{MSF} are described.

\paragraph{Setup}
The two experiments are performed on two separate walking excavators. For both machines the sensor suit described in Section~\ref{sec:implementation_details} is used. The first machine is equipped with an \textit{Ellipse-A} IMU unit at the bottom of the machine cabin, while the second makes use of a \textit{Lord MicroStrain MV5} IMU sensor mounted on the roof of the machine. Note that other than the extrinsic sensor calibration all experiments are run with the same (noise) parameters. For both settings the IMU measurements are available at $100$Hz, GNSS at $20$Hz, and the lidar odometry runs at $5$Hz.

\paragraph{Outline}
To fulfill the requirements of autonomous operation in complex environments, high global accuracy, smooth local consistency, and low-latency update rates are required.
All three criteria are thoroughly evaluated in situations with and without GNSS reception in the following sections. A video of the conducted experiments and the derived results is submitted along with this publication.

\begin{figure}[t]
\includegraphics[width=\columnwidth]{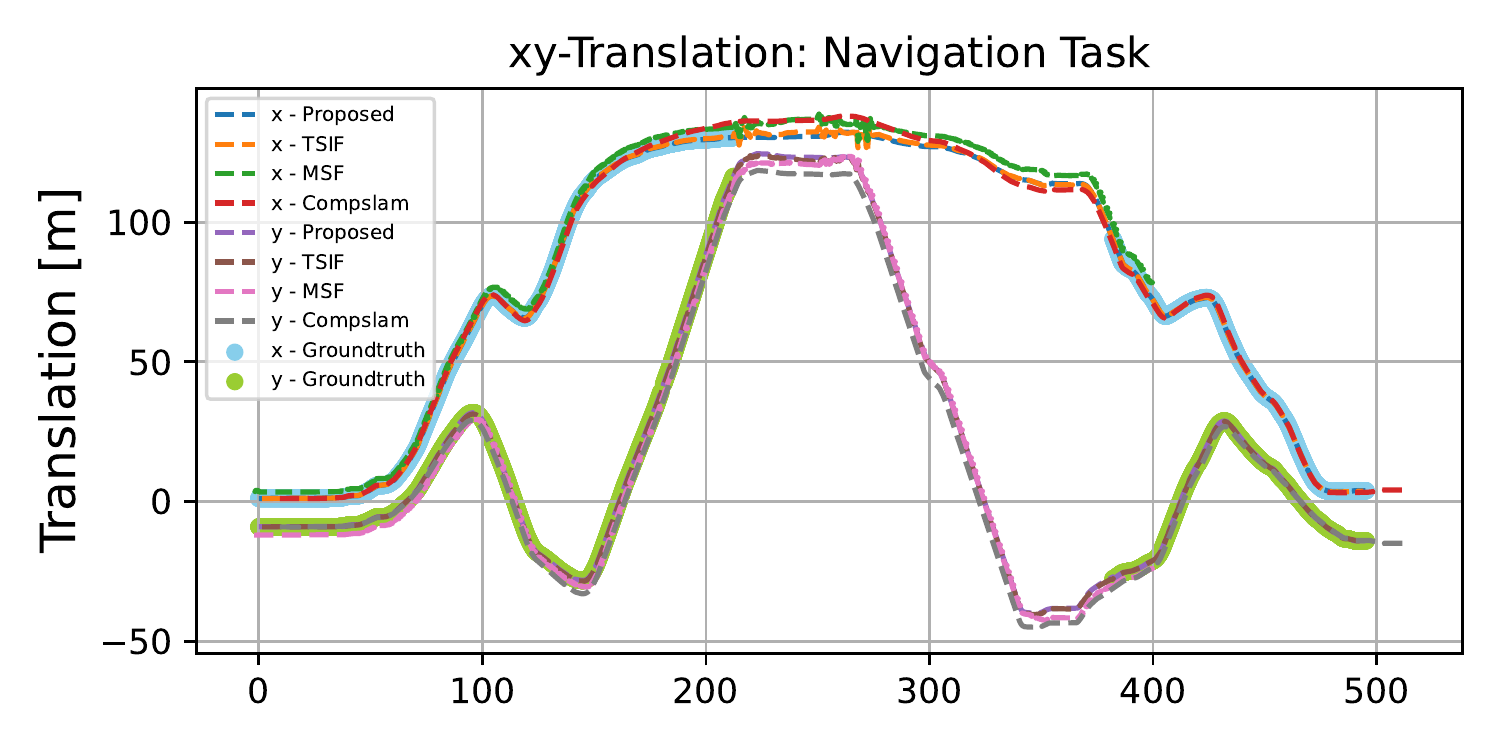}
\caption{
Global accuracy evaluation for the navigation task. Global accuracy is maintained as compared to CompSLAM, while local smoothness is also maintained in absence of GNSS. Time is expressed in [s].}
\vspace{-2ex}
\label{fig:accuracy_nav_task}
\end{figure}

\begin{figure*}[h!]
\centering
\includegraphics[width=1.0\textwidth]{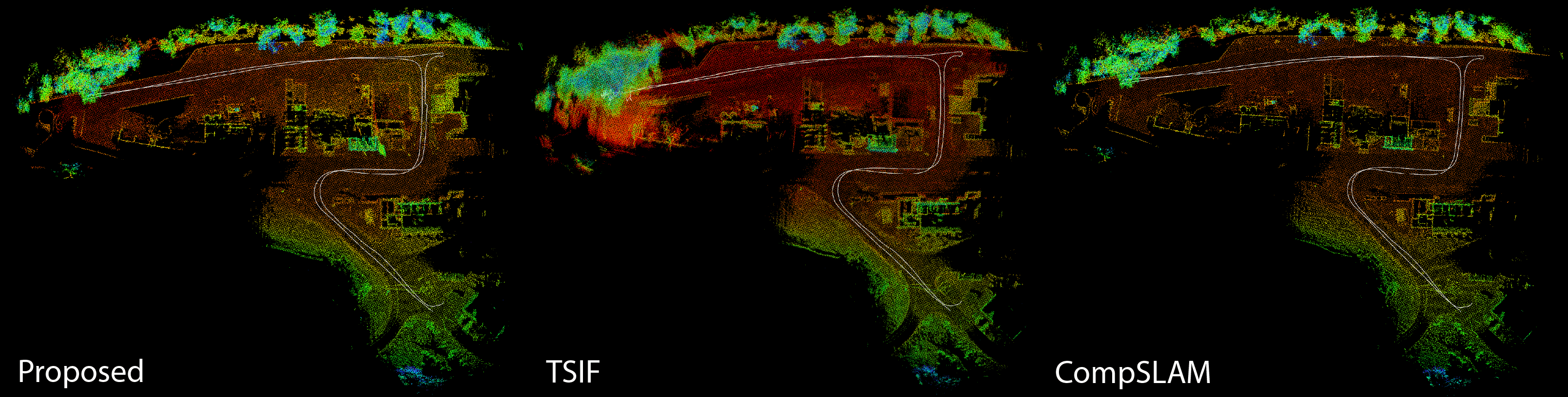}
\caption{Maps that are built using the estimates from the proposed approach, TSIF, and CompSLAM (left to right). While the TSIF map gets corrupted close to trees, the proposed approach makes use of the estimates provided by the lidar odometry during these intervals.}
\label{fig:experiment1_maps}
\vspace{-2ex}
\end{figure*}

\subsection{Global Accuracy}

Global accuracy of the proposed approach is evaluated in two ways: first, the robot trajectories for both experiments are compared against ground-truth, MSF and TSIF, and second, the estimates are used for building a map of the environment.

\subsubsection{Global Localization}
First, during the construction task all the three approaches are compared against the position estimates of the RTK GNSS, which is considered as groundtruth for fully functional situations.
In order to understand the position accuracy, the upper plot in Figure~\ref{fig:accuracy} shows the relative position error for the three approaches with quantitative results presented in Table~\ref{table_delta_prediction}. It can be seen that the proposed approach demonstrates similar sub-centimeter level accuracy and achieves the smallest standard deviation for its low-latency state-estimates.
\begin{table}[h!]
\vspace{-1ex}
    \centering
    \caption{Relative Position Errors for Construction Task}
    \begin{tabular}{cc|cc|cc} 
           \toprule
          \multicolumn{2}{c}{Proposed} & \multicolumn{2}{c}{TSIF} & \multicolumn{2}{c}{MSF} \\ 
         \midrule
          $\mu$ & $\sigma$ & $\mu$ & $\sigma$ & $\mu$ & $\sigma$ \\ 
          $0.18$\,\si{\cm} & $1.28$\,\si{\cm} & $0.20$\,\si{\cm} & $1.54$\,\si{\cm} & $0.10$\,\si{\cm} & $1.60$\,\si{\cm} \\
            \bottomrule
    \end{tabular}
    \label{table_delta_prediction}
    \vspace{-1ex}
\end{table}
In order to show the consistency of the proposed approach during periods of sensor dropout, the translation along Z-axis for the same dataset is investigated more closely and shown in the lower plot of Figure~\ref{fig:accuracy}. During the interval from $105$ to $140$s seconds the GNSS signal is considered unreliable due to a high covariance provided by the Leica system, and hence, is not used as an input to the fallback graph. By making full use of the lidar measurements during that time, the presented approach remains as smooth as the CompSLAM while preserving global accuracy. In contrast, TSIF performs well during intervals with GNSS available but becomes unreliable in its absence. MSF remains operational as it integrates both GNSS and lidar measurements, however, a drift is introduced due to sensor dropout and recovery.

For the navigation task, a similar evaluation is performed and shown in Figure~\ref{fig:accuracy_nav_task}. During this experiment, as the robot followed a long trajectory along X and Y axes with little change along Z-axis, the $(x,y)$ coordinates are shown for plot clarity. A similar behaviour can be observed; while being operated in the neighborhood of trees between second $220$ and $380$, GNSS accuracy is degraded. TSIF and MSF show perturbation due to GNSS loss while the proposed approach and CompSLAM estimates remain smooth. Note that latter accumulates drift in contrast to the proposed approach.

\subsubsection{Mapping Task}
Next, the suitability of the state estimates for building large-scale maps in real-time is investigated. Three maps are shown in Figure~\ref{fig:experiment1_maps}. From left to right the following estimates were used for building the map; first the output of the proposed state estimator is used, in the second image the map built using the output of TSIF is shown, while the most right plot is the internal map of CompSLAM. It can be seen that in the top left part of the image the quality of the map built with TSIF is corrupted due to GNSS loss because of partial coverage with trees. The map built using estimates of the proposed approach looks similar to CompSLAM map, while being globally accurate, demonstrating the benefits of fusing multiple modalities for the considered type of tasks.

\subsection{State Estimation}
The pose estimation consistency of the proposed approach for low-latency state-estimation is evaluated in this section.
\subsubsection{Consistency}
\begin{figure}[h!]
\adjustbox{trim=0 {0.08\height} 0 0,clip, width=\columnwidth}
{\includegraphics[width=\columnwidth]{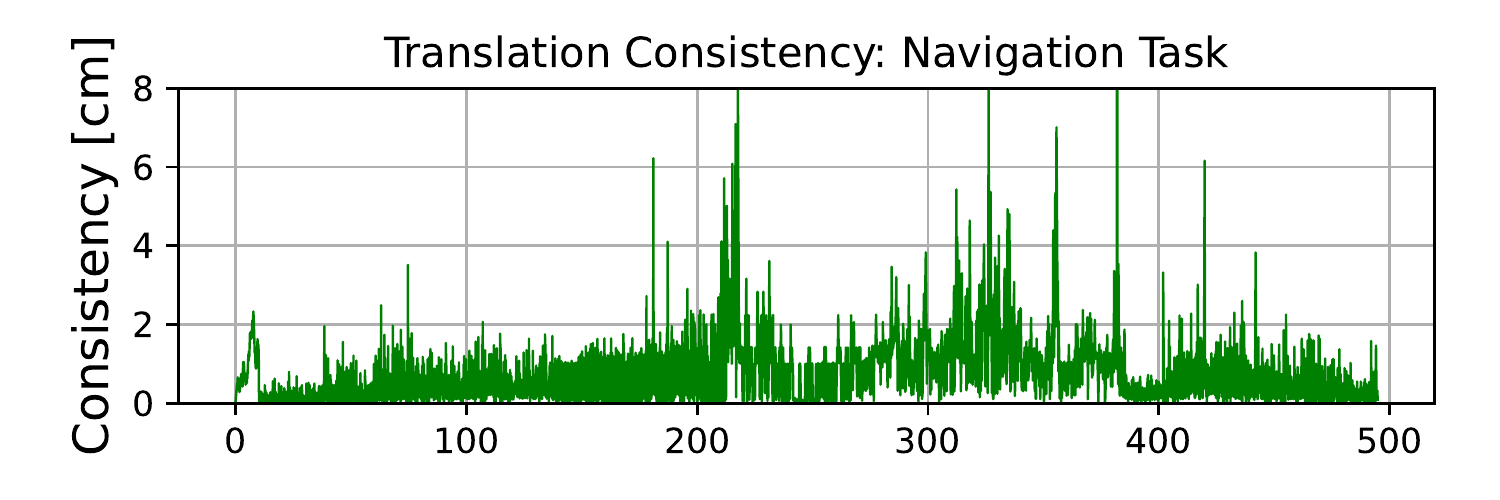}}
\caption{State-estimation consistency results for the navigation task compute but calculating the deviation between propagated and optimized estimates. In this plot, time is expressed in [s].}
\label{fig:consistency-arche}
\vspace{-2ex}
\end{figure}

To evaluate the consistency of the proposed approach an analysis similar to the one proposed in~\cite{sandy2019confusion} is performed. The consistency is given by computing the deviation between high update rate propagated poses and the updated state estimates (after an optimization of the considered states is performed). Evaluation of translation consistency is presented in Figure~\ref{fig:consistency-arche} for the navigation task. While the prediction estimates are provided at $100$Hz, the optimized estimates arrive in at roughly $25$Hz, as they are triggered by either GNSS or lidar measurements. It should be noted that when dealing with large machines precise extrinsic calibrations between sensors are difficult to obtain, which can  deteriorate the performance of sensor-fusion based estimates, however, the proposed approach achieves a consistency error with a mean of $0.58$\,\si{\cm} and a standard deviation of $3.16$\,\si{\cm}.

\subsubsection{Latency}
Finally, the state-estimator latency is investigated and compared to TSIF and MSF, which are both filtering-based approaches whose main advantage is considered to be a low latency.
As shown in Table~\ref{table:latency}, the latency - defined by the mean and standard deviation of the time that passes between the arrival of an IMU measurement until the propagated state-estimate is available - is not only real-time capable, but even outperforms the latency of TSIF, and is comparable to the latency of MSF. 
\begin{table}[h!]
    \vspace{-2ex}
    \centering
    \caption{State-estimation latency during the navigation task.}
    \begin{tabular}{cc|cc|cc} 
         \toprule
          \multicolumn{2}{c}{Proposed} & \multicolumn{2}{c}{TSIF} & \multicolumn{2}{c}{MSF} \\ 
          \midrule
          $\mu$ & $\sigma$ & $\mu$ & $\sigma$ & $\mu$ & $\sigma$ \\ 
          $41 \mu$s & $15 \mu$s & $102 \mu$s & $40 \mu$s & $32.9 \mu$s & $18.7 \mu$s \\
          \bottomrule
    \end{tabular}
    \label{table:latency}
    \vspace{-2ex}
\end{table}

\section{Conclusions}\label{sec:conclusions}
In this paper a multi-sensor fusion framework based on factor graphs was proposed to provide reliable and accurate pose estimation for construction excavators. The proposed approach combines the benefits of filtering and smoothing methods to enable fast and consistent state estimation for control while providing accurate localization for global large-scale operations and mapping. Real-world tests conducted on two excavators demonstrate the reliability of the proposed approach by handling cases of sensor dropout and recovery that occur naturally during excavator operations. Future work will focus on joint estimation of chassis orientation, encoder bias, extrinsic calibrations and sensor time-offsets. 


\bibliographystyle{IEEEtran}
\bibliography{ICRA2022}

\end{document}